\documentclass{article}

    \usepackage[final]{neurips_2020}

\usepackage[utf8]{inputenc} 
\usepackage{graphicx}
\usepackage{subfigure}

\usepackage{changepage}

\usepackage[T1]{fontenc}    
\usepackage{hyperref}       
\usepackage{url}            
\usepackage{booktabs}       
\usepackage{amsfonts}       
\usepackage{nicefrac}       
\usepackage{microtype}      
\usepackage{biblatex}

\usepackage{hyperref}
\hypersetup{colorlinks,linkcolor={blue},citecolor={blue},urlcolor={red}}
\title{A Novel Approach for Semiconductor Etching Process\\
with Inductive Biases}

\author{%
  Sanghoon Myung\thanks{Primary author, shoon.myung@samsung.com} \\
  Data and Information Technology (DIT) Center\\
  Samsung Electronics\\
  \And
  Hyunjae Jang \\
  DIT Center \\
  Samsung Electronics\\
  \And
  Byungseon Choi \\
  DIT Center \\
  Samsung Electronics\\
   \And
  Jisu Ryu \\
  DIT Center \\
  Samsung Electronics\\
   \And
  Hyuk Kim \\
  Semiconductor R\&D Center \\
  Samsung Electronics\\
   \And
  Sang Wuk Park \\
  Semiconductor R\&D Center \\
  Samsung Electronics\\
  \And
  Changwook Jeong\thanks{Corresponding autho, chris.jeong@samsung.com} \\
  DIT center\\
  Samsung Electronics\\
   \And
  Dae Sin Kim \\
  DIT Center \\
  Samsung Electronics\\
}

\begin{document}

\maketitle

\begin{abstract}
  The etching process is one of the most important processes in semiconductor manufacturing. We have introduced the state-of-the-art deep learning model to predict the etching profiles. However, the significant problems violating physics have been found through various techniques such as explainable artificial intelligence and representation of prediction uncertainty. To address this problem, this paper presents a novel approach to apply the inductive biases for etching process. We demonstrate that our approach fits the measurement faster than physical simulator while following the physical behavior. Our approach would bring a new opportunity for better etching process with higher accuracy and lower cost.

\end{abstract}

\section{Introduction}

Artificial intelligence (AI) has undergone a renaissance recently, and it exceeds human performance based on a number of data and its low cost in various domains such as vision and language. These success stories have lured various big data companies in the manufacturing industry (e.g. semiconductor) to attempt to introduce the deep learning model. Nevertheless, it is still hard to find such stories in the real world. One of the major reasons is that in the test phase, the data distribution is often shifted. For instance, distributional shift is caused by high dispersion between production equipment in the manufacturing. In this case, the conventional deep learning model still does not reach human performance; it even predicts the absurd results. To tackle this problem, there are three approaches. First, prediction uncertainty is investigated by previous works\cite{gal2016dropout, lakshminarayanan2017simple}. It is crucial that the model can represent whether the results are trustworthy or not, which also helps the decision making. The second approach is explainable AI (xAI) to interpret how the model makes the decision\cite{lundberg2017unified,ribeiro2016should,vaswani2017attention}. The most of previous works introduce the new interpretable model to explain the deep learning with proper assumptions or insert the interpretable layer (i.e. attention layer) in the neural network. When xAI results are well matched with the domain knowledge, this approach can give confidence to human. If not, even the wrong results can inspire humans to modify the model. The third approach is to apply the inductive biases, which helps the model predict outside of training-range data. The first and second methods are most preferred in terms of universality. These two methods help humans to make decisions, however, they cannot improve accuracy level. The third method, inductive biases, may improve accurate prediction, but it needs to be specifically designed for each problem. This paper, with semiconductor etching process, investigates the problem of data shift based on xAI and uncertainty. It also suggests a novel approach with proper inductive biases for etching process.

\section{Methodology for Plasma Etching}

We will briefly introduce the semiconductor dataset for etching. The most devices of semiconductor enforce electrical isolation to prevent leakage currents. One of the ways to impose electrical isolation is called the Deep Trench Isolation (DTI)\cite{rung1982deep}. It is essential to optimize the recipe to etch the DTI shape we design. But there are a number of controllable parameters to develop etching process in a heuristic and iterative manner. Hence, the engineers rely on the Technology Computer-aided Design (TCAD) simulators to optimize the recipe with lower cost than experiment. However, TCAD simulators require proper calibration of the model parameters to match the simulation results with the measurements. In addition, TCAD approach may be inaccessible for some cases due to its extremely long turn-around-time (TAT). At last, even if these problems are definitely resolved, it is difficult to accurately design the DTI because the physical models cannot capture all the relevant features in experiment. If deep learning can replace TCAD, the development costs can be significantly reduced because of its faster speed to infer the result. To add more details, Section 2.1 briefly reviews the background of semiconductor to understand the etching problem with TCAD simulation. Section 2.2 describes the baseline model to predict the DTI shape and defines the problem from interpretation of the baseline model. Section 2.3 suggests the inductive biases to resolve the problem described in Section 2.2.

\subsection{TCAD simulation}

TCAD simulation is performed using Synopsys Sentaurus 3D topography\cite{synopsys2016sentaurus}. In order to predict how deep the silicon is etched, TCAD simulation solves the complex phenomena based on physical equations. Figure 1 (a) shows the etching recipe example of the inputs of the model to etch DTI, whereas Figure 1 (b) describes that of the outputs measured by destroying inspection. Figure 1 (c) illustrates the physical phenomena in which the neutrons and ions, separated by plasma gas, react. As discussed above, TCAD simulators need lots of calibration parameters, thus they consume high TAT. Another problem of TCAD simulation is that parameters are calibrated with equipment by equipment due to equipment dispersion.

\begin{figure}[h]
    \centering
    \includegraphics[width=0.8\textwidth]{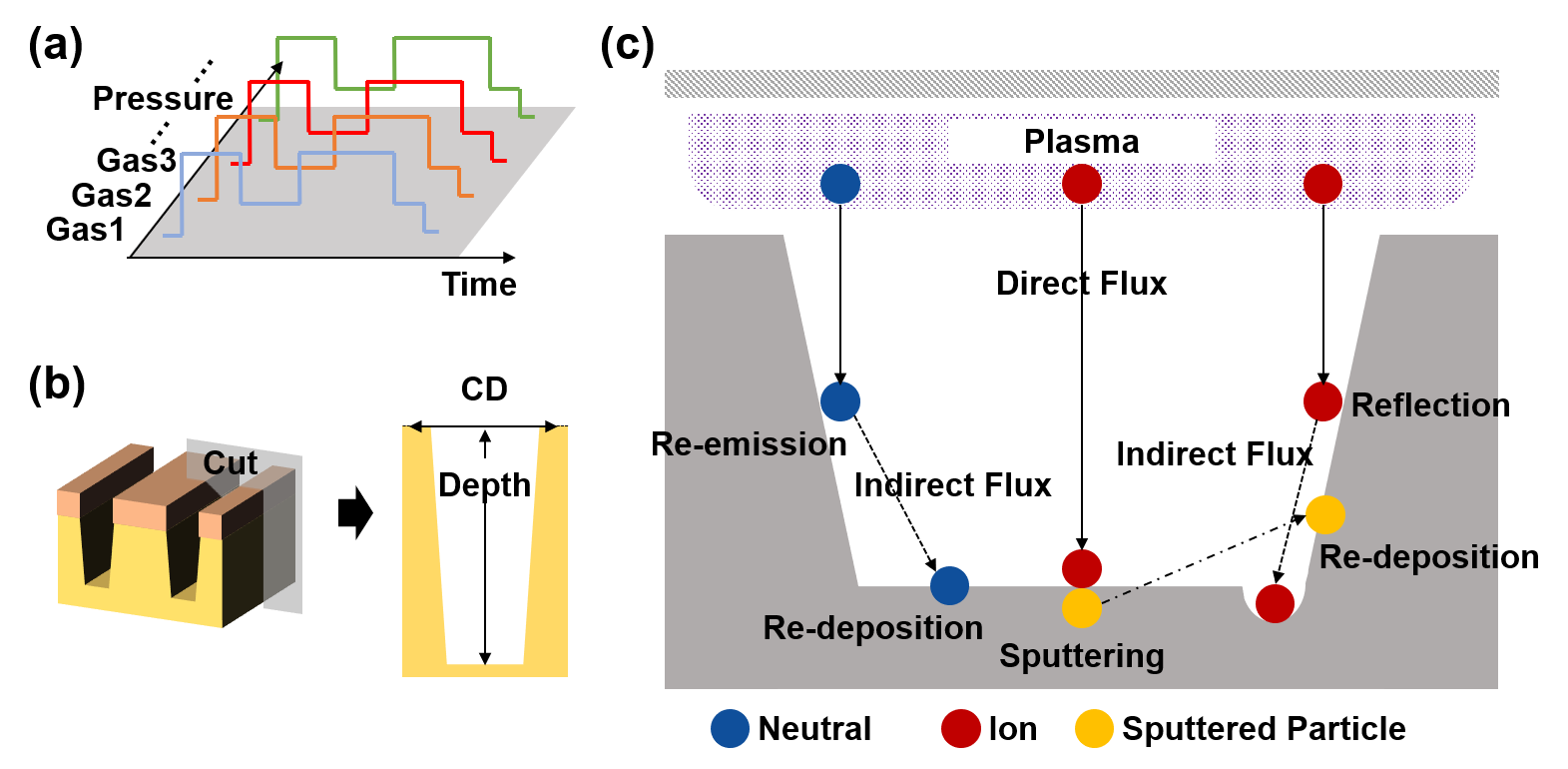}
    \caption{(a) Description of the input recipe of TCAD simulation (b) Description of the output (c) A summary of physical effects on etching.\centering
    }
\end{figure}

\subsection{Baseline model and its interpretation}

Since both inputs and outputs can deal with sequential data, we use the Transformer\cite{vaswani2017attention} model. Categorical inputs, which are not available for TCAD simulation (e.g. equipment information, wafer location), are also used as our model inputs. We design that the model generates the DTI profile as input recipes and then, adapts its shape from categorical inputs. Finally, we introduce a deep ensemble method based on Negative-log-likelihood (NLL) loss to measure the prediction uncertainty. Now, we firstly inspect how proper our baseline model works. An attention map shows that all steps have contributed to etch the top of DTI (Fig. 2 (a)). On the other hand, only last step has affected the bottom of DTI (Fig. 2. (b)). These results are a good agreement with domain knowledge. Second, we experiment how model etches the trench per each step. Interestingly, the model predicts that the bottom is etched by the first step (Fig. 3 (a)) out of our expectation. This result is regarded as a significant fault even considering the prediction uncertainty. The model also occurs additional errors in which the trench is covered again even after being etched by the previous step.

\begin{figure}[t] 
\minipage{0.45\linewidth}
  
  \includegraphics[width=\textwidth]{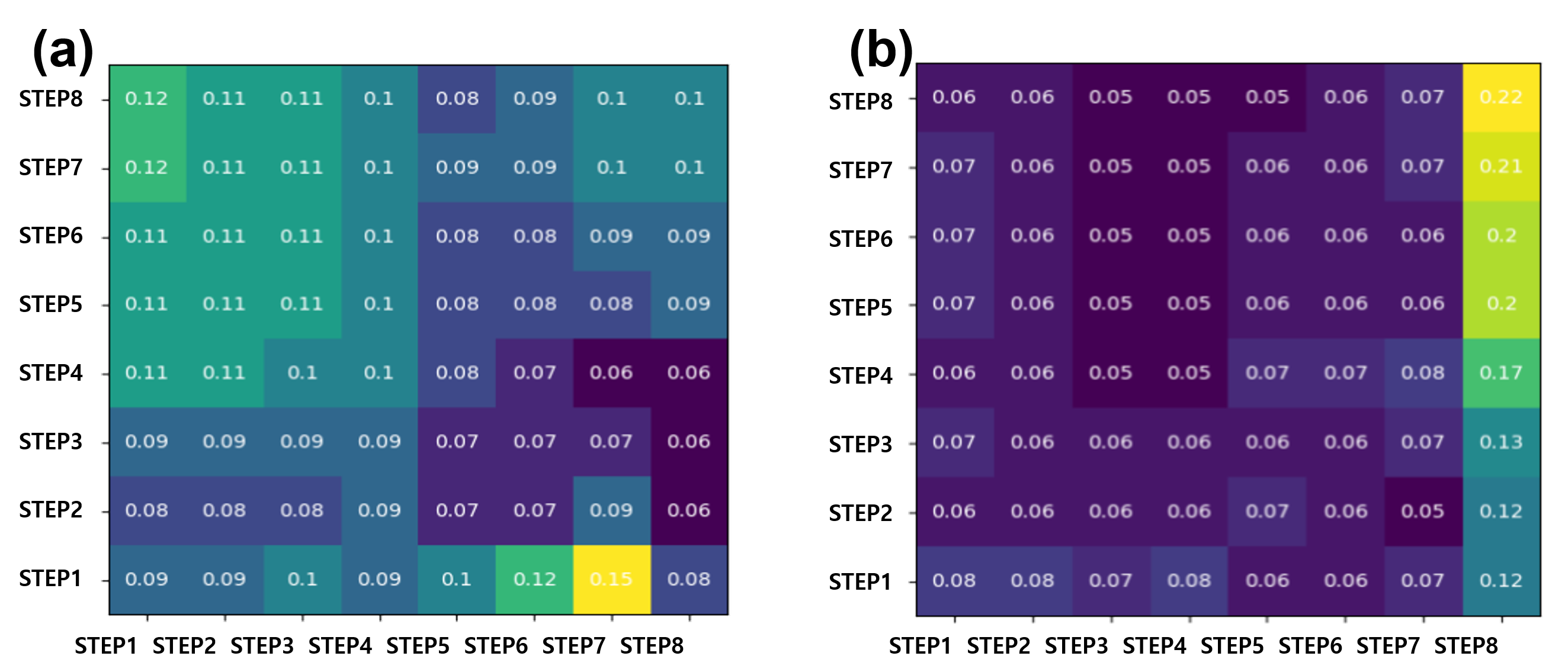}
  \caption{The attention plots of model when etching (a) the top, (b) the bottom.
}
\endminipage\hfill\centering
\minipage{0.45\linewidth}
  \includegraphics[width=\textwidth]{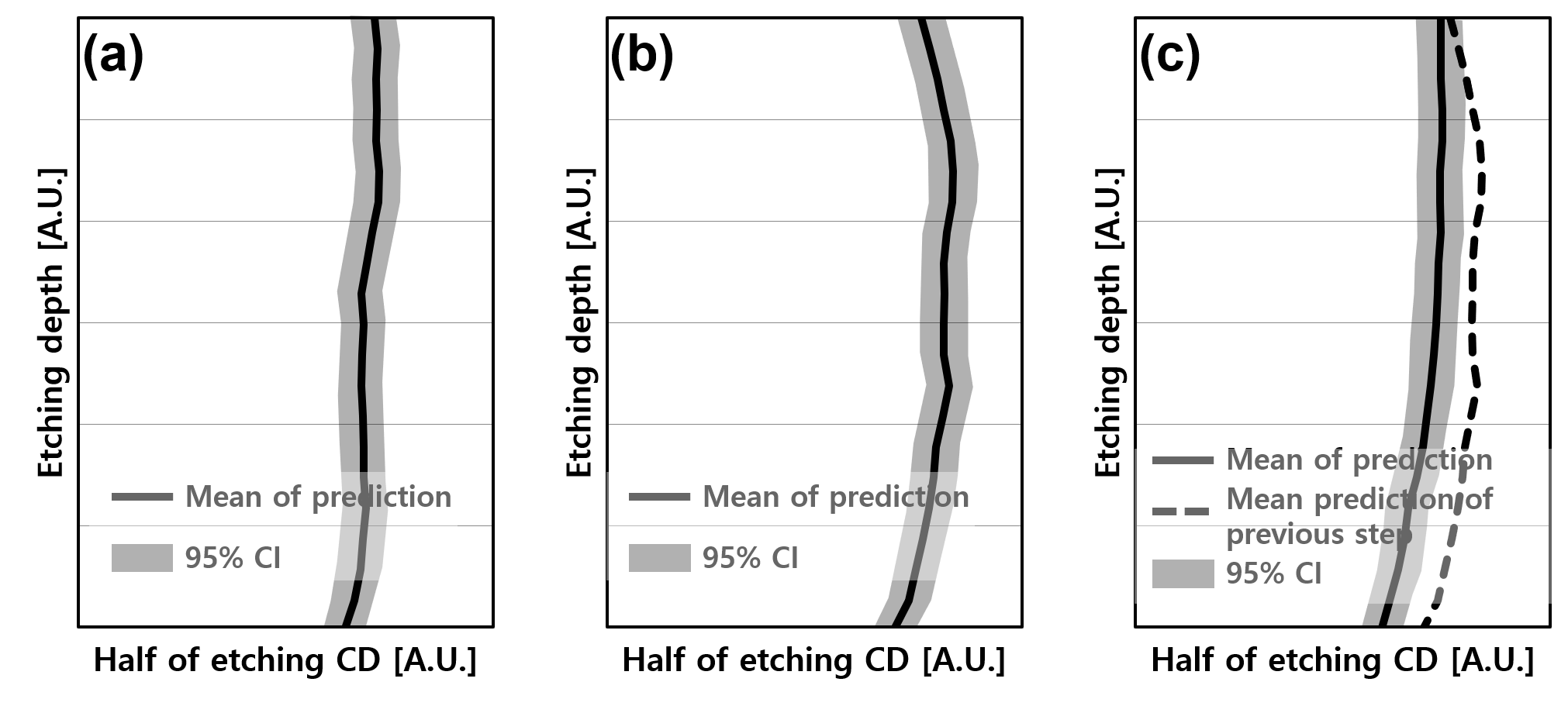}
  \caption{The prediction results at the (a) first (b) immediate (c) final step.
}
\endminipage\hfill\centering

\end{figure}

\subsection{Inductive biases for the etching model}

Although the data distribution is shifted in the test phase, the problem described in Section 2.2 is caused by inference based on distribution of training samples. There are two main ideas to be introduced based on the analysis in Section 2.2. \\
\begin{adjustwidth}{1em}{1em}
1) DTI is etched horizontally first, and then etched vertically as the step progresses. To simply realize this phenomena, we hypothesize that the etched DTI profile per each step should follow the exponential function such as Weibull distribution. The model generates k and $\lambda$ in each input recipe step and then predicts the DTI shape as the k and $\lambda$ parameters pass through the Weibull activation. One of the advantages of this assumption is that two parameters control the DTI profile. Figure 4 (a) shows ideal results of etching profile based on the domain knowledge.\\
\\
2) The silicon is always etched as steps advance. To implement this phenomena into the baseline model, the output of the network is designed to forecast the amount of etching at the moment of the step. The amount of the etched DTI is accumulated in the final step in order to prevent the DTI from being covered. Figure 4 (b) depicts the DTI shape in each step.\\
\end{adjustwidth}

Finally, we summarize the model architecture in Fig. 4 (c).
\begin{figure}[h]
    \centering
    \includegraphics[width=\textwidth]{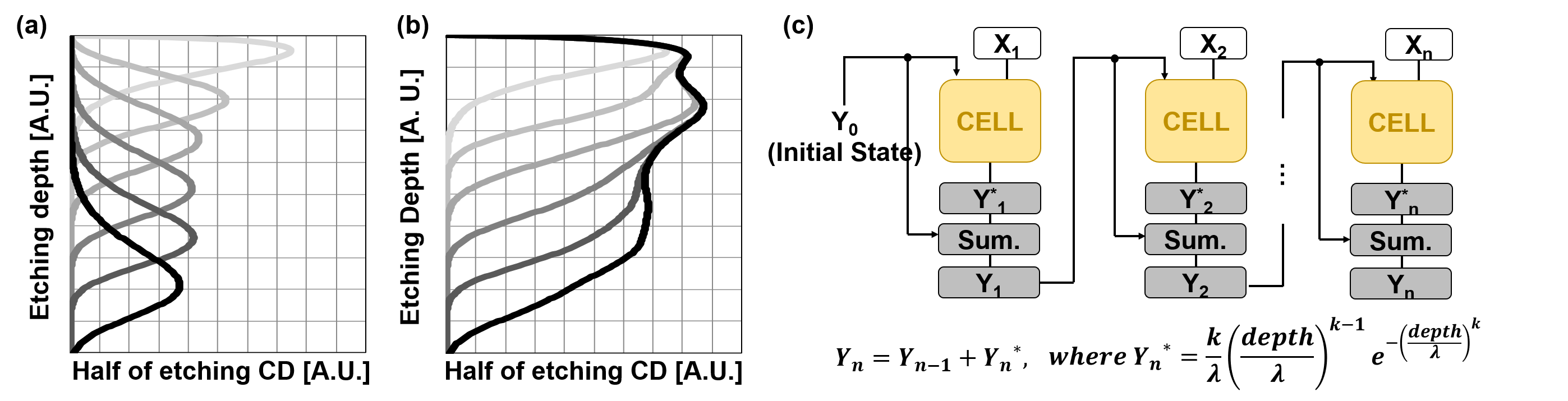}
    \caption{(a) The etching profile per each step. (b) The accumulated etching profile as the step progresses. (c) Illustration of the model architecture.\centering
}
\end{figure}

\section{Experimental Results and Discussion}

To assess the effectiveness of our assumption, the prediction of our model should be compared with the real DTI profile at each step; however, they are not available due to substantial cost in development phase. Therefore, to validate the effectiveness, we experiment in two folds. We inspect how much prediction results of the model and TCAD simulations are similar in Section 3.1. We also test the generalization ability of the model compared with the baseline model in Section 3.2. At the end, we discuss the results.

\subsection{Ablation study}

Figure 5 shows the prediction results of TCAD simulation. As described in Section 2.2, it is confirmed that the DTI is vertically etched as the step progresses. However, the final shape derived from TCAD simulation would not match with actual process results because of the following reasons. 1) The physical models to describe the etching process are a work in progress. 2) TCAD simulators require proper calibration of the model parameters to match the measurements; though, it would be difficult to fit the measurements because there exist numerous calibration parameters. Figure 5 (b) interprets our model prediction results applied to inductive biases. It is validated to be a good agreement with physics of TCAD simulation. It is much better matched with the measurement of the final step than TCAD simulation results. If Weibull activation is not applied, the DTI is not properly etched even after the model learns the fact that the DTI is always etched (Fig. 5(c)). This result is much different from TCAD simulation.

\subsection{Generalization test}

We assess how our model works better than baseline model. The training data and test data are recipes consisted of five to nine steps and ten to eleven steps, respectively (Fig. 6 (a)). Our model’s score has been improved by 9.4\% compared to the baseline model (Fig. 6 (b)). The reason why the improvement is lower than expectation is that output distribution is not changed although input distribution is shifted. Nevertheless, NLL score is much better than the baseline model because the prediction uncertainty is accumulated after each step (Fig. 7).

\begin{figure}[h] 
\minipage{0.5\linewidth}
  
  \includegraphics[width=\textwidth]{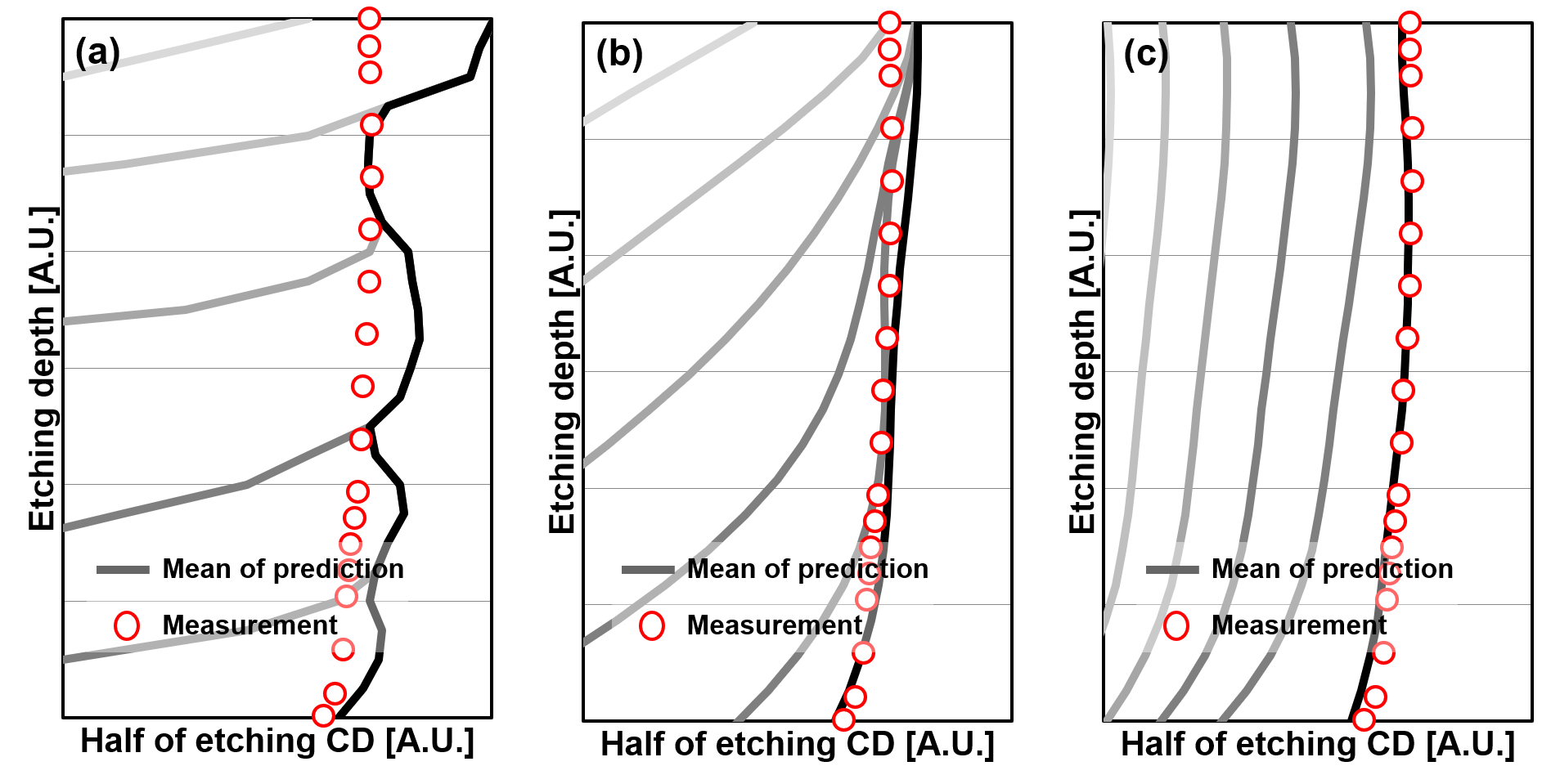}
  \caption{Prediction results.
\centering}
\endminipage\hfill\centering
\minipage{0.5\linewidth}

  \includegraphics[width=\textwidth]{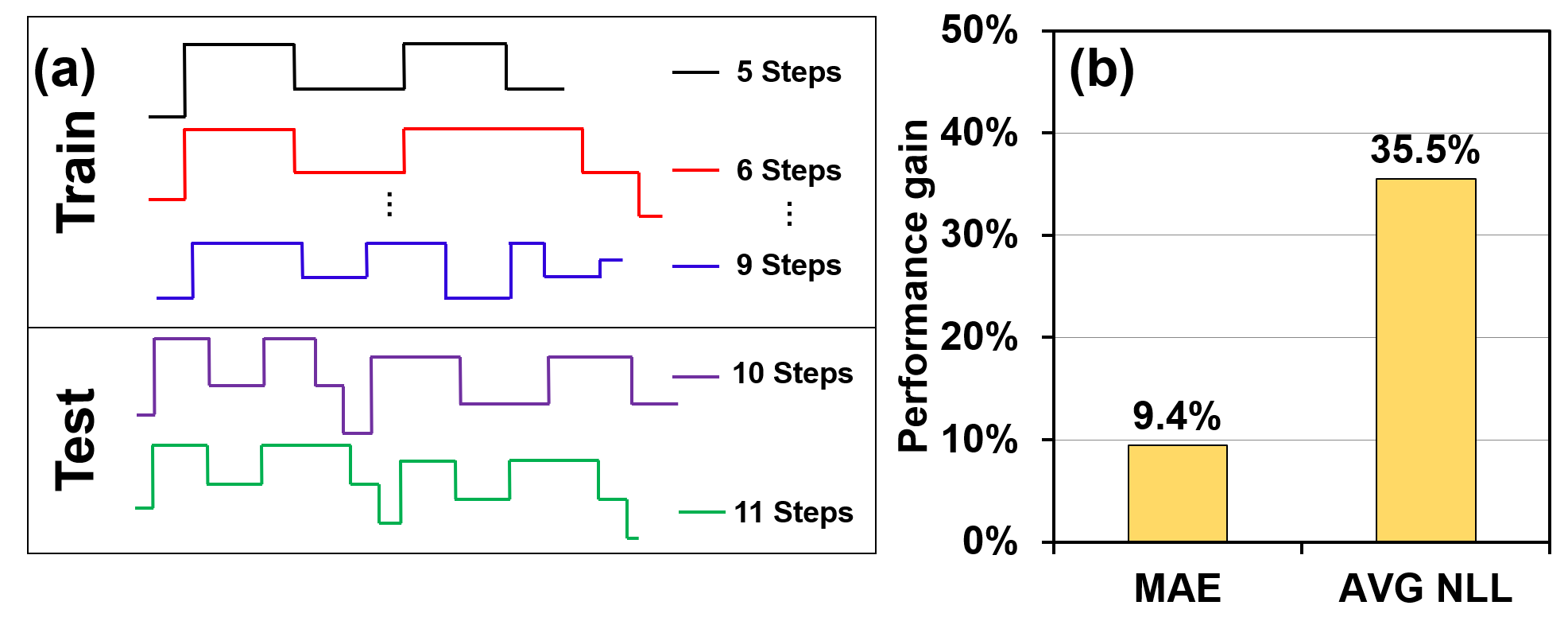}
  \caption{(a) Explanation of training and test input recipe (b) Result of generalization test.
}
\endminipage\hfill\centering

\end{figure}

\begin{figure}[h]
    \centering
    \includegraphics[width=\textwidth]{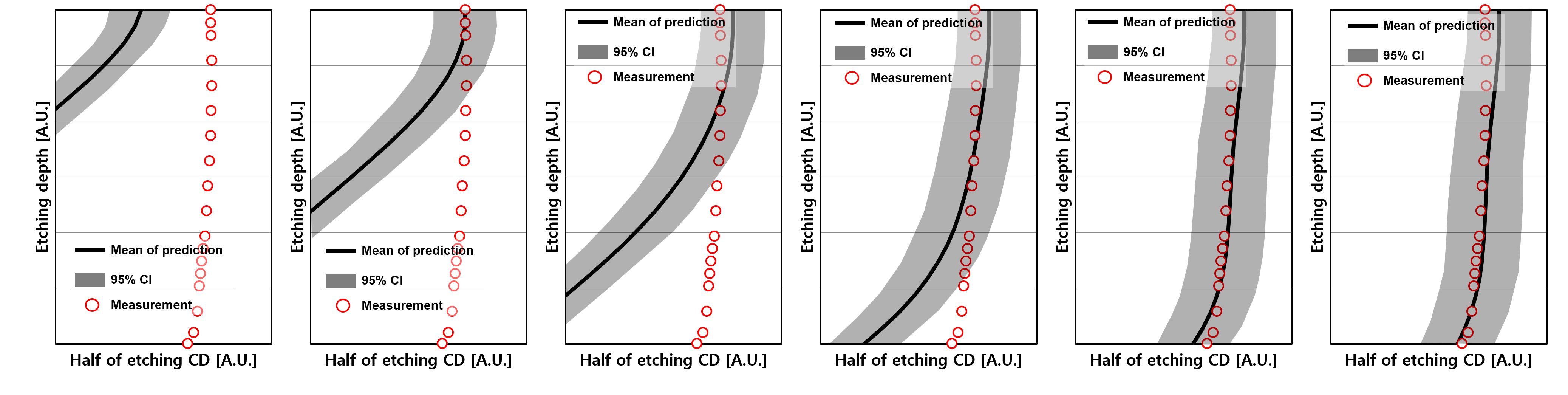}
    \caption{Prediction results with the uncertainty as the each step advances.
\centering
}
\end{figure}
\section{Conclusion}

We present a novel approach applying the inductive biases for etching process to learn causality between each step. Our approach can generate profile at any step without intermediate data. Our model proves that the DTI shape is a good agreement with TCAD simulation results. Above all, our model can easily fit the measurement faster than TCAD simulation. These results may open possibilities for a new, more convenient and more accurate way than TCAD simulation.

\printbibliography

\end{document}